\pdfoutput=1
\documentclass[11pt]{article}

\usepackage[]{EMNLP2023}

\usepackage{times}
\usepackage{latexsym}
\usepackage{times}
\usepackage{latexsym}
\usepackage{url}
\usepackage[T1]{fontenc}
\usepackage{multirow}
\usepackage{graphicx}
\usepackage{hyperref}
\usepackage{hhline}
\usepackage{amsmath}
\usepackage{caption}
\usepackage{subcaption}
\usepackage[T1]{fontenc}
\usepackage[utf8]{inputenc}

\usepackage{microtype}

\usepackage{inconsolata}

\title{ChiSCor: A Corpus of Freely Told Fantasy Stories by Dutch Children for Computational Linguistics and Cognitive Science}

\author{Bram van Dijk*\textsuperscript{1}, Max van Duijn*\textsuperscript{1}, Suzan Verberne\textsuperscript{1}, \and Marco Spruit\textsuperscript{1,}\textsuperscript{2}\\
 \textsuperscript{1}Leiden Institute of
  Advanced Computer Science \\
  \textsuperscript{2}Leiden University Medical Centre \\
  \texttt{\{b.m.a.van.dijk, m.j.van.duijn, s.verberne, m.r.spruit\}} \\
  \texttt{@liacs.leidenuniv.nl}}

\begin{document}
\maketitle

\def\thefootnote{*}\footnotetext{Equal contribution.}\def\thefootnote{\arabic{footnote}}

\begin{abstract}
%1. What did we do
In this resource paper we release ChiSCor, a new corpus containing 619 fantasy stories, told freely by 442 Dutch children aged 4-12.
%2. Why did we do it
ChiSCor was compiled for studying how children render character perspectives, and unravelling language and cognition in development, with computational tools. Unlike existing resources, ChiSCor's stories were produced in natural contexts, in line with recent calls for more ecologically valid datasets.
%3. How did we do it
ChiSCor hosts text, audio, and annotations for character complexity and linguistic complexity. Additional metadata (e.g. education of caregivers) is available for one third of the Dutch children. ChiSCor also includes a small set of 62 English stories. 
%4. What did we find
This paper details how ChiSCor was compiled and shows its potential for future work with three brief case studies: i) we show that the syntactic complexity of stories is strikingly stable across children's ages; ii) we extend work on Zipfian distributions in free speech and show that ChiSCor obeys Zipf's law closely, reflecting its social context; iii) we show that even though ChiSCor is relatively small, the corpus is rich enough to train informative lemma vectors that allow us to analyse children's language use. We end with a reflection on the value of narrative datasets in computational linguistics.

\end{abstract}

\section{Introduction}
All of us tell stories on a daily basis: to share experiences, contextualise emotions, exchange jokes, and so on. There is a rich tradition of research into how such storytelling develops during infancy, and its relations with various aspects of children's linguistic and cognitive development \citep[for an overview see][]{cremin2016storytelling}. ChiSCor (\textbf{Chi}ldren's \textbf{S}tory \textbf{Cor}pus) was compiled to give a unique impulse to this tradition: it allows for (computationally) studying how children render character perspectives such as perceptions, emotions, and mental states throughout their cognitive and linguistic development.
 
Existing research connecting language and cognition has largely relied on standardised tests \citep[for review see][]{milligan2007language}. Yet, recently researchers across fields have urged for data reflecting phenomena they study in their natural context. For instance, computational linguists call for better-curated and more representative language datasets \citep{bender2021dangers, PAULLADA2021100336}, language pathologists question whether standardised linguistic tests capture children's actual linguistic skills \citep{ebert2014relationships}, and cognitive scientists call for more naturalistic measures of socio-cognitive competences \citep{beauchamp2017neuropsychology, nicolopoulou2017narrativity, rubio2021pragmatic}. Following these considerations, ChiSCor has three key features: it contains fantasy stories that were told \textit{freely}, within children's \textit{social} classroom environments, and stories are supplemented with relevant \textit{metadata}. As such, ChiSCor documents a low-resource language phenomenon, i.e. freely produced and socially embedded child language.

This paper makes the following contributions. First, we release ChiSCor and describe its compilation, data, and annotations in detail (Sections \ref{Background} and \ref{Corpus_compilation}). Second, we show how ChiSCor fuels future work on the intersection of language, cognition, and computation, with three brief case studies (Section \ref{case_studies}). We explore the Dependency Length Minimization hypothesis \citep{futrell2015large} with ChiSCor's language features and show that the syntactic complexity in children's stories is strikingly stable over different age groups. Also, we extend emerging work on Zipf's law in speech \citep[e.g.][]{lavi2023zipfCDS,linders2023zipfrevisited} and find that ChiSCor's token distribution approximates Zipf better than a reference corpus consisting of language written by children, which we explain by appealing to the Principle of Least Effort. Furthermore, we show that ChiSCor as a small corpus is rich enough to be used with NLP-tools traditionally thought to require large datasets. We train informative lemma vectors with ChiSCor, that can be used to analyse how coherently children use specific lemmas of interest, and potential bias in their language use.

Together, our case studies demonstrate that even though storytelling is a cognitively challenging task, the language children employ is no less sophisticated \citep[an observation also supported by][]{dijk2021modelling, van-dijk-etal-2023-theory}. And although corpora of narratives are often smaller, we show that we can (and should) leverage NLP-tools to unravel linguistic and cognitive mechanisms at work in children's language productions. As discussed in Section \ref{Discussion}, we see this as an important stepping stone towards building more ecologically valid language models.

\section{Background and relevance}\label{Background}
Various resources of Dutch child language exist. Before the 2000s, corpora typically consisted of child speech gathered in unstructured home settings involving smaller numbers of younger children \citep[e.g.][]{schlichting1996discovering, wijnen1998acquisition}. Later, more structured language elicitation (e.g. with picture books) from larger samples of children was more common \citep[e.g.][]{kuijper2015he}, and recently we have seen large corpora documenting thousands of essays in school settings \citep{tellings2018basiscript}, and many hours of speech recordings in human-machine interaction contexts \citep{cucchiarini2013jasmin}. 

Although these resources are valuable, what is currently lacking is a corpus of speech samples that are i) produced freely in natural social settings, while being ii) sufficiently independent or `decontextualised' to be a good reflection of children's capacities, and iii) containing metadata about children's backgrounds. The rest of this section will discuss these three characteristics, on the basis of which ChiSCor was compiled.

\begin{table*}[ht]
\centering 
\small {\begin{tabular}{c c c}
  \textbf{Type} & \textbf{Quantity} & \textbf{Details} \\ [0.5ex] 
  \hline
  Audio & \textasciitilde11.5 hours & 619 44.1kHz .wav files \\
  Text & 619 stories & \textasciitilde74k words, verbatim and normalized .txt files \\
  Metadata & All 442 children & School grade (reflecting age group) \\ 
  Extra metadata & 148 children & Exact age, reading time, education parents, no. of siblings, \\
  & & gender, lang. disorder (y/n), home language Dutch (y/n) \\
  Linguistic features & All 619 stories & E.g. vocabulary perplexity, vocabulary diversity, syntactic tree depth, \\ 
  & & words before root verb, syntactic dependency distance  \\
  Annotations & All 619 stories & Character complexity (see Section \ref{annotations_vars}) \\
  \hline
\end{tabular}}
\caption{Details on ChiSCor's data. Besides the Dutch stories, ChiSCor also features an additional set of 62 English stories, for which audio, text, (extra) metadata, linguistic features and annotations are also available.} 
\label{tab:1}
\end{table*}

\begin{table*}[ht]
\begin{center}
\small {\begin{tabular}{|| c | c | c ||}
    \hline
    \textbf{Level} & \textbf{Example} & \textbf{ID}  \\ [0.5ex]
    \hline
    \hline
    \multirow{3}{*}{Actor}
    & \textit{Once upon a time there was a castle.} & \\
    & \textit{There stood a throne in the castle and \underline{a princess sat on the throne}.} & 093101 \\
    & \textit{And the princess had a unicorn.} & \\
    \hline
    \multirow{4}{*}{Agent}
    & \textit{Once upon a time there as a prince and he saw a villain.} & \\
    & \textit{\textit{\underline{And then he called the police.}}} & 023101 \\
    & \textit{And then the police came.} & \\
    & \textit{And then he was caught. The end.} & \\
    \hline
    \multirow{4}{*}{Person}
    & \textit{Once upon a time there was a girl.} & \\
    & \textit{\underline{She really wanted to play outside}. Her mother did not allow it.} & 010101 \\
    & \textit{She went outside anyway and her mother asked where are you going?} & \\ 
    & \textit{And the girl said I am going outside. The end.} & \\
    \hline
\end{tabular}}
\end{center}
\caption{Translated stories from ChiSCor, traceable with ID. Underscoring shows the character the label is based on.}
\label{tab:2}
\end{table*}

\textbf{i)} The stories in ChiSCor were collected on a large scale in natural settings, because language as a social phenomenon is highly context-sensitive. The corpora mentioned above that include such settings are often limited in scale, whereas the newer corpora are large-scale, but cover language produced for a machine interface or in school assignment context, thus are not socially embedded.

\textbf{ii)} The stories in ChiSCor concern a special form of \textit{decontextualized} language use, in which children cannot draw on cues (like picture books), feedback from interlocutors (as they could in a conversation), or much shared background knowledge with the audience (that hears a new fantasy story). Thus, the cognitive demands in producing decontextualized language are high, since children have to simultaneously plan the story, monitor their language use, and make sure the audience can follow the plot \citep{nicolopoulou2019using}. As such, eliciting freely-told narratives is an acknowledged method for sampling an individual child's language skills on phonological, lexical, syntactic, and pragmatic levels \citep{southwood2004comparison, ebert2014relationships, nicolopoulou2015using}, as well as for assessing cognitive abilities, including memorizing, planning, organizing world knowledge \citep{mckeough2003transformation}, and Theory of Mind \citep{nicolopoulou1993play}. Furthermore, proficiency in decontextualized language is known to be a good predictor of literacy and academic achievement \citep{snow1991some}. As far as we know, no larger-scale corpora of decontextualized Dutch child speech exist, and in the international context such corpora are also rare.
    
\textbf{iii)} Existing resources often contain data on children's age and gender, but not on their backgrounds such as the educational levels of parents, which ChiSCor does contain (see Section \ref{Corpus_compilation}). Metadata on subjects included in datasets becomes increasingly important, e.g. for gauging how representative language samples are \citep{bender2021dangers}, but also for follow-up work where e.g. partitioning the dataset is desired.

\section{Corpus compilation}\label{Corpus_compilation}

\subsection{Data collection}
We contacted primary schools, a day care and a community center in the South and South-West of The Netherlands to offer storytelling workshops, in the period 2020-2023. Workshops generally consisted of three stages: first, we openly brainstormed with children about what stories are, without enforcing our own ideas (e.g. what is a story, where can you find stories, what do you like about stories); second, we invited children to freely fill in the details of a fantasy story initiated by us as experimenters (e.g. filling in names, settings, events in a variation on the King Midas avarice myth); third and most importantly, we challenged children to individually make up and tell a fantasy story to their class peers, which we recorded.

Our storytelling workshop was inspired by the Story Telling Story Acting (STSA) paradigm, originally developed by \citet{paley1990boy} and used as a framework in empirical studies by \citet{nicolopoulou2007actors}, \citet{nicolopoulou2015using} and \citet{NICOLOPOULOUAgeliki2022Atha}. Work by Nicolopoulou generally targets younger children using a longitudinal research practice integrated in the school curriculum, which involves both telling stories and acting them out. Our approach differs in that we included all primary school age groups (4-12y), but focused on storytelling only. Like in the STSA paradigm, children told stories live to an audience of peers, which comes close to narration in everyday social life: children explored themes like friendship and conflict, excitement over real and imagined events, and storytelling was interactive in the sense that their class peers reacted with laughter, disbelief, and so on. 

High-quality recordings were made with a Zoom H5 recorder. Recordings were manually transcribed into verbatim and normalised versions. In the normalised stories employed in the case studies (Section \ref{case_studies}), noise such as false starts and broken-off words was manually corrected with as little impact on semantics and syntax as possible. Our project was approved by the Leiden University Science Ethics Committee (ref. 2021-18). Caregivers were informed beforehand and could optionally provide additional metadata, which \textasciitilde33\% (148) did. Our corpus, metadata, and code are available on OSF.\footnote{\url{https://shorturl.at/bvGOX}.} See for more details on the data Table \ref{tab:1} and for sample stories Table \ref{tab:2}.

\subsection{Metadata}\label{metadata}
\begin{figure}[t]
\includegraphics[width=\linewidth]{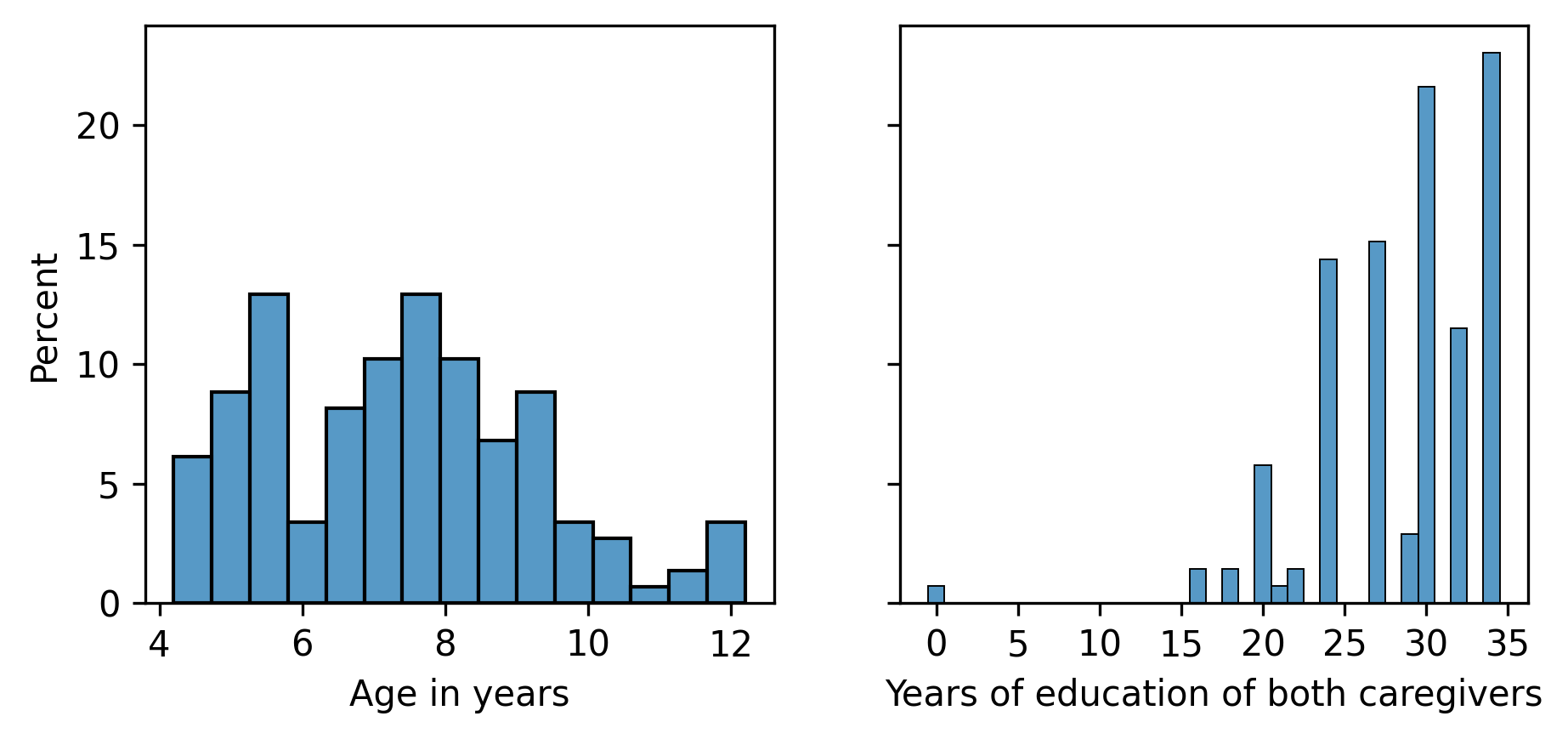}
\caption{Ages of 148 children and educational levels of their caregivers. Bars in each plot stack up to 100\%.}
\label{fig:1}
\end{figure}

Here we highlight two variables from the metadata we collected: children's age and the educational levels of caregivers. Most ages are well-represented (Figure \ref{fig:1}), but older children (ages 10-12) are under-represented; less teachers from older age groups signed up for the workshop. For educational levels, we see that \textasciitilde53\% of the children has two highly educated caregivers (in the Dutch system, a higher degree equals a minimum of 15 years of education), while \textasciitilde24\% has caregivers with two vocational (or lower) degrees (a vocational degree equals a maximum of 12 years of education) \citep{CPB2012education}. Thus, in the part of our sample for which extra metadata is available, children from caregivers with higher socioeconomic status (SES) are over-represented. Yet, selection bias is higher in the metadata than in the language samples in ChiSCor as a whole: while we were able to include stories told by children from schools in more challenged neighbourhoods in ChiSCor, metadata depended on caregivers filling out forms, which caregivers with higher SES did more often.

\subsection{Annotations}\label{annotations_vars}
Here we highlight two types of annotations available in ChiSCor: socio-cognitive annotations in the form of character complexity annotations, and linguistic annotations in the form of automatically extracted features.

Regarding \textbf{social cognition}, ChiSCor provides character complexity annotations that involve one label per story indicating the `depth' of the most complex character encountered in a story (examples in Table \ref{tab:2}). Character depth can be used as a window into the socio-cognitive skills of storytellers and was adapted from \citet{nicolopoulou2007actors} and \citet{nicolopoulou2016promoting}. The scale ranges from `flat' \textit{Actors} merely undergoing or performing simple actions, to \textit{Agents} having basic perceptive, emotional, and intentional capacities, possibly in response to their environments, to `fully-blown' \textit{Persons} with (complex) intentional states that are explicitly coordinated with the storyworld. Labelling was done with CATMA 6 \citep{evelyn_gius_2021_5015305} and in-text annotations are available on OSF. Labelling character depth requires expert annotation, given that children's stories often progress in non-obvious ways. Interrater agreement was obtained in two rounds. Two experts A and B first labelled a random subset of 8\% of stories, yielding moderate agreement (Cohen's $\kappa$ = .62). After calibration (discussing disagreements to consensus), A labelled the rest of the corpus, and B labelled another random 8\% as second check, for which Cohen's $\kappa$ = .84 was obtained, indicating almost perfect agreement \citep{landis1977measurement}.

Regarding \textbf{linguistic features}, we extracted mean dependency distance between syntactic heads and dependents as measure of syntactic complexity with spaCy 3.5 \citep{honnibal-johnson:2015:EMNLP}. We follow \citet{liu2008dependency} and \citet{liu2017dependency} and calculated mean dependency distance with $ DD(S) = \frac{1}{n-s}\sum_{i=1}^n \left | DD_i \right | $, where $DD_i$ is the absolute distance in number of words for the $i$-th dependency link, $s$ the number of sentences and $n$ the number of words in a story. Language employing larger dependency distances is more demanding for working memory, thus harder to process \cite{grodner2005consequences, futrell2015large}. We further elaborate on dependency distance in a case study in Section \ref{Syntactic_comp}. 

We emphasise that many more linguistic features are included on OSF than we can discuss here, e.g. lexical perplexity and syntactic tree depth as common measures of linguistic proficiency and development \citep[e.g.][]{mcnamara2014automated, kyle2016measuring, dijk2021modelling}.

\section{Case studies with ChiSCor}\label{case_studies}
We conduct three small case studies to illustrate ChiSCor's potential. Since we aim to show ChiSCor's versatility to the broader community, we draw in Study 1 (Section \ref{Syntactic_comp}) on ChiSCor's own linguistic annotations and metadata; in Study 2 leverage ChiSCor in a corpus linguistics-style analysis on Zipf's law in child speech (Section \ref{Zipf}), and in Study 3 show the feasibility of using ChiSCor with NLP-tools that are traditionally thought to require larger corpora (Section \ref{W2V}).

\subsection{Case study 1: Syntactic Complexity} \label{Syntactic_comp}
\begin{figure}[h]
\includegraphics[width=\linewidth]{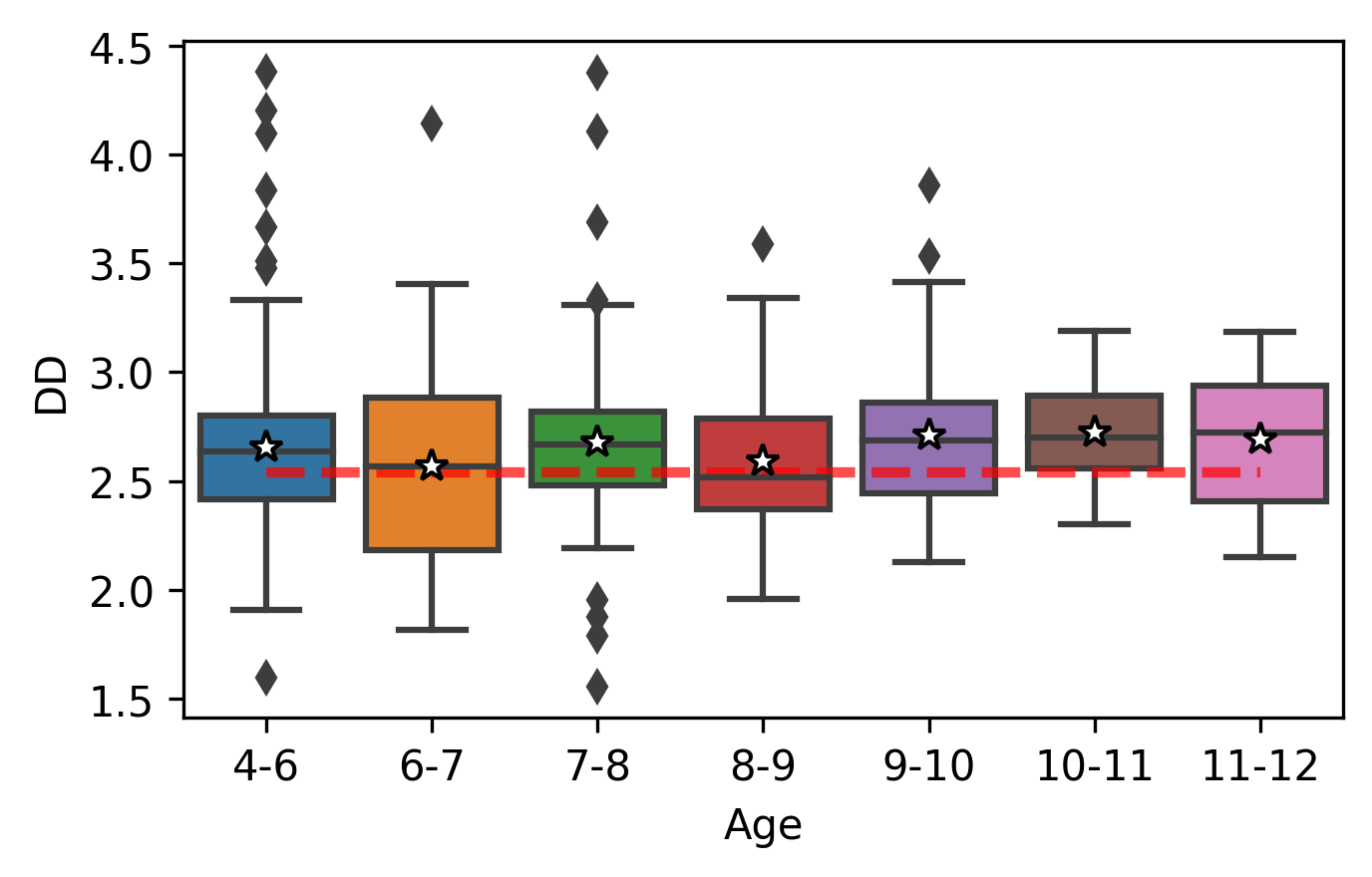}
\caption{Dependency Distance (DD) conditioned on age groups as customary in Dutch primary education. Dashed line indicates mean DD reported by \citet{liu2008dependency}. Stars indicate means.}
\label{fig:dd}
\end{figure}

The Dependency Length Minimization (DLM) hypothesis states that languages have evolved to keep syntactically related heads and dependents close together (such as an article modifying a noun), so that anticipation of a noun after an article is not stretched over many intervening words, which increases cognitive load and/or working memory costs \citep{futrell2015large}. Although DLM has been observed for various languages in various studies \citep[e.g.][]{gildea2010grammars,futrell2015large}, as far as we know, DLM for child speech has not been explored. ChiSCor concerns live storytelling, which is known to be a cognitively intense language phenomenon (see Section \ref{Background}), which makes the DLM interesting to explore in ChiSCor's context. It is intuitive to expect that children employ smaller dependency distances to reduce cognitive load. We leverage ChiSCor's linguistic features (dependency distance as explained in Section \ref{annotations_vars}) and metadata (age groups) to analyse the developmental trend under the DLM. Especially for younger children (e.g. 4-6y), DLM could be expected to be more pronounced, given that they are arguably less proficient language users with little formal language training in school.  
Our modelling approach was as follows. In a linear model we included contrast-coded predictors, such that each predictor indicated the mean dependency distance difference with the previous grade (`backwards difference coding'), to model a trend over age groups. Dependency distance conditioned on age is plotted in Figure \ref{fig:dd} for 442 stories of 442 children, and coefficients of the model are given in Table \ref{tab:3}. Note that for those children who told multiple stories, we included only the first story to maximize independence of observations.

\begin{table}[h]
\centering 
\small {\begin{tabular}{c c c c}
  \textbf{Predictor} & $\beta$ & SE & \textit{p} \\ [0.5ex] 
  \hline
  \textit{Intercept}&  2.66 & .02 &  .00 \\
  Diff. 6-7/4-6     &  -.09 & .07 &  .20 \\
  Diff. 7-8/6-7     &   .11 & .07 &  .13 \\
  Diff. 8-9/7-8     &  -.09 & .06 &  .16 \\ 
  Diff. 9-10/8-9    &   .12 & .07 &  .08  \\ 
  Diff. 10-11/9-10  &   .01 & .10 &  .91  \\
  Diff. 11-12/10-11 &  -.03 & .12 &  .81 \\
  \hline
\end{tabular}}
\caption{Coefficients of the linear model. Each predictor indicates the difference in DD with the previous age group.} 
\label{tab:3}
\end{table}

Dependency distance appeared to be surprisingly stable across age groups: no single predictor significantly predicted dependency distance (Table \ref{tab:3}, all $ p > .05$), nor did all predictors together $(F_{6, 435} = 1.078, p = .38, R^2_{adj} < .01$). Contrary to expectations, it was not the case that younger children, as less proficient language users, employ shorter dependency distances, nor do children employ longer dependency distances as they grow older. Interestingly, in backwards difference coding, the intercept is the grand mean of dependency distance of all groups (2.66), which is close to the mean dependency distance (2.52) found for Dutch written by adults and reported by \citet{liu2008dependency}. 

\begin{figure}[t]
\includegraphics[width=\linewidth]{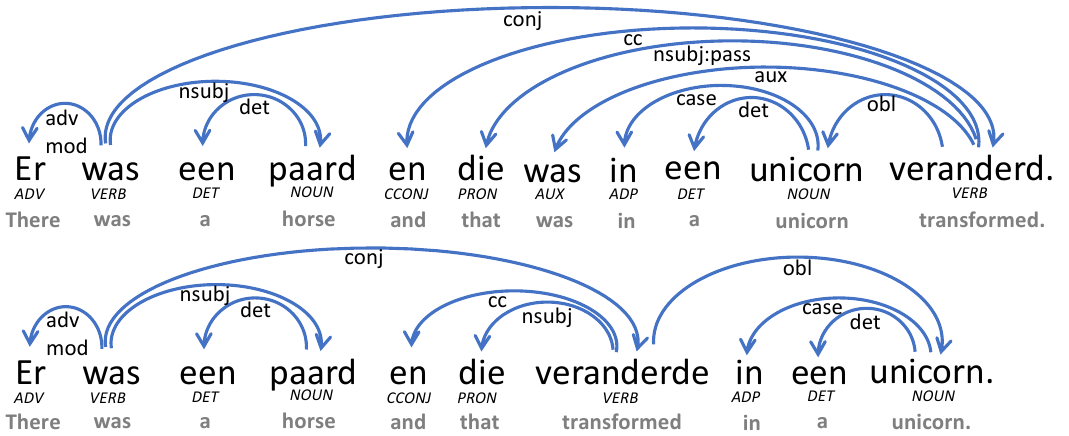}
\caption{Top: original utterance from story 033201 in PaP with mean dep. dist. = 3.2. Bottom: paraphrase in SP (bottom) with mean dep. dist. = 2.}
\label{fig:pap}
\end{figure}

We make a start with trying to explain why, in storytelling for younger children (4-6y), we find higher dependency distances than expected. Manual examination of narratives from this group showed that children often use syntactically complex constructions to refer to past events, even when simpler alternatives are available or preferred. The typical tense for narrative contexts is the Simple Past (SP) for many languages \citep{zeman2016perspectivization}, and SP can be used for completed and ongoing events in the past \citep{boogaart1999aspect} in the storyworld. SP is syntactically simple; it requires only a single inflected verb. Young children, however, often use Present/Past Perfect (PrP/PaP) and Past Progressive (PP) constructions. These forms are used to indicate ongoing (PrP/PP) and completed (PaP) events in the past, and are syntactically similar in that they all involve an auxiliary depending on a (past) participle (PrP/PaP) or infinitive (PP) that is typically at utterance-final position, thus creating complex syntax. Figure \ref{fig:pap} provides an illustration from our data of a child narrating a completed past event in PaP, which pushes dependency distance well beyond the average reported by \citet{liu2008dependency}, although the more efficient option would be SP.

Although it is known that young children in experimental contexts also refer to past events with PrP and PP constructions instead of SP \citep{schaerlaekens1993taalverwerving, van2010gaan}, in the context of decontextualized language use and the DLM our finding was unexpected. We find a possible explanation in work by \citet{van2010gaan}: separating tense (auxiliary) from lexical information (verb) yields more complex syntax on the one hand, but makes processing easier for an audience on the other hand. After all, the audience does not have to decode different types of information packed in a single inflected verb. The trade-off between syntactic simplicity and ease of processing could indeed explain why ChiSCor's spoken narratives, produced live in front of an audience of peers, contain relatively high proportions of PrP and PP. Follow-up work would be needed to further substantiate this idea.

\subsection{Case study 2: Zipf distributions}\label{Zipf}
Zipf distributions, where token frequencies are proportional to their rank $r$ according to $f(r)\propto{\frac{1}{r^\alpha}}$ with $\alpha=1$ \citep{zipf1932selected}, were found for many language samples \citep{xiao2008applicability, ferrer2005variation, Yu2018Zipf, Smith2007Investigation, tellings2014basilex, lavi2023zipfCDS}, but are also subject to debate \citep[for review see][]{piantadosi2014zipf}; is Zipf a trivial mathematical artefact or a fundamental property of human cognition and language? As \citet{linders2023zipfrevisited} note, to answer this question we should analyze Zipf in more natural forms of communication, such as speech instead of written language, and invoke cognitive mechanisms underlying Zipf, such as the Principle of Least Effort (PLE). The PLE assumes that senders prefer efficient communication using infrequent, hence often shorter and ambiguous words, whereas receivers prefer larger vocabularies of longer, infrequent words to more easily decode messages. Zipf distributions are considered the balanced trade-off between sender and receiver needs \citep{cancho2003least}. 

The PLE is salient in ChiSCor's context: since live storytelling is a cognitively intense form of decontextualized language use (Section \ref{Background}), this could lead to a bias in storytellers towards frequent tokens, to alleviate cognitive load, a prediction made by \citet{linders2023zipfrevisited}. Yet, at the same time, if receiver needs are neglected, they cannot follow along; receivers cannot ask for clarification during storytelling as would be possible in e.g. normal conversations, which is something senders take into account to prevent losing their audience, which equals losing the point of storytelling. This balance is arguably less pronounced in written discourse, where there is opportunity to reconsider earlier parts, and no immediate interaction, thus less pressing receiver needs. Here we pit the token distribution of ChiSCor against that of BasiScript, a corpus of \textit{written} child language (subsection `free essays', \textasciitilde3.4M tokens from thousands of Dutch children of 7-12 year \citep{tellings2018basiscript}), to compare Zipfian distributions in speech to the written domain.

We followed \citet{piantadosi2014zipf} in performing a binomial split on the observed frequency of each token to avoid estimating frequency and rank on the same sample. We used Zipf's original formula introduced above rather than derivations to model token distributions, following  \citet{linders2023zipfrevisited}. We log-transformed (base 10) token rank and frequency to model Zipf linearly with $log(frequency) = log(intercept) + slope * log(rank)$. 

\begin{figure}[t]
\includegraphics[width=\linewidth]{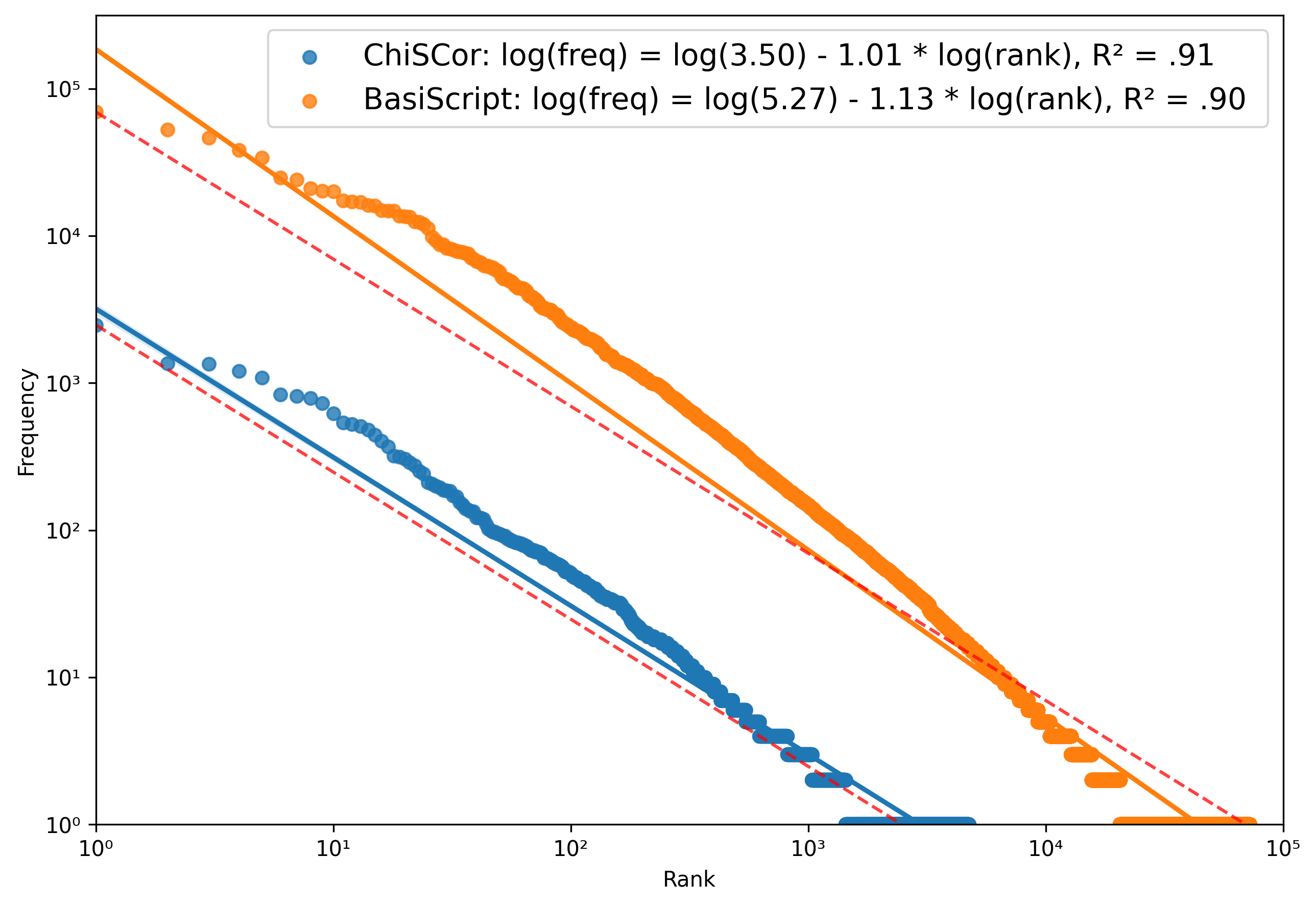}
\caption{Rank-frequency plots of ChiSCor and BasiScript. Dashed lines indicate Zipf's law with $\alpha=1$, blue/orange lines indicate model fits.}
\label{fig:zipf}
\end{figure}

We see in Figure \ref{fig:zipf} that both corpora approximate the plotted Zipf lines with good model fits ($R^2\ge.90$). Yet, ChiSCor approximates the Zipf line more closely than BasiScript, with a slope closer to $-1$, supporting the idea that in live storytelling, balancing sender \textit{and} receiver needs is more pressing than in written language, even though in live storytelling a bias towards frequent tokens seems intuitive. The larger negative slope (-1.13) fitted for BasiScript indicates that senders rely more on frequent tokens and employ less infrequent tokens, which confirms the prediction that in written discourse, receiver needs are less pressing. Follow-up work could investigate Zipf distributions in both corpora beyond tokens, e.g. on parts-of-speech or utterance segments \citep{lavi2023zipfCDS, linders2023zipfrevisited}.

\vfill

\begin{figure*}{}
\centering
    \begin{subfigure}[b]{0.5\textwidth}            
            \includegraphics[width=\textwidth]{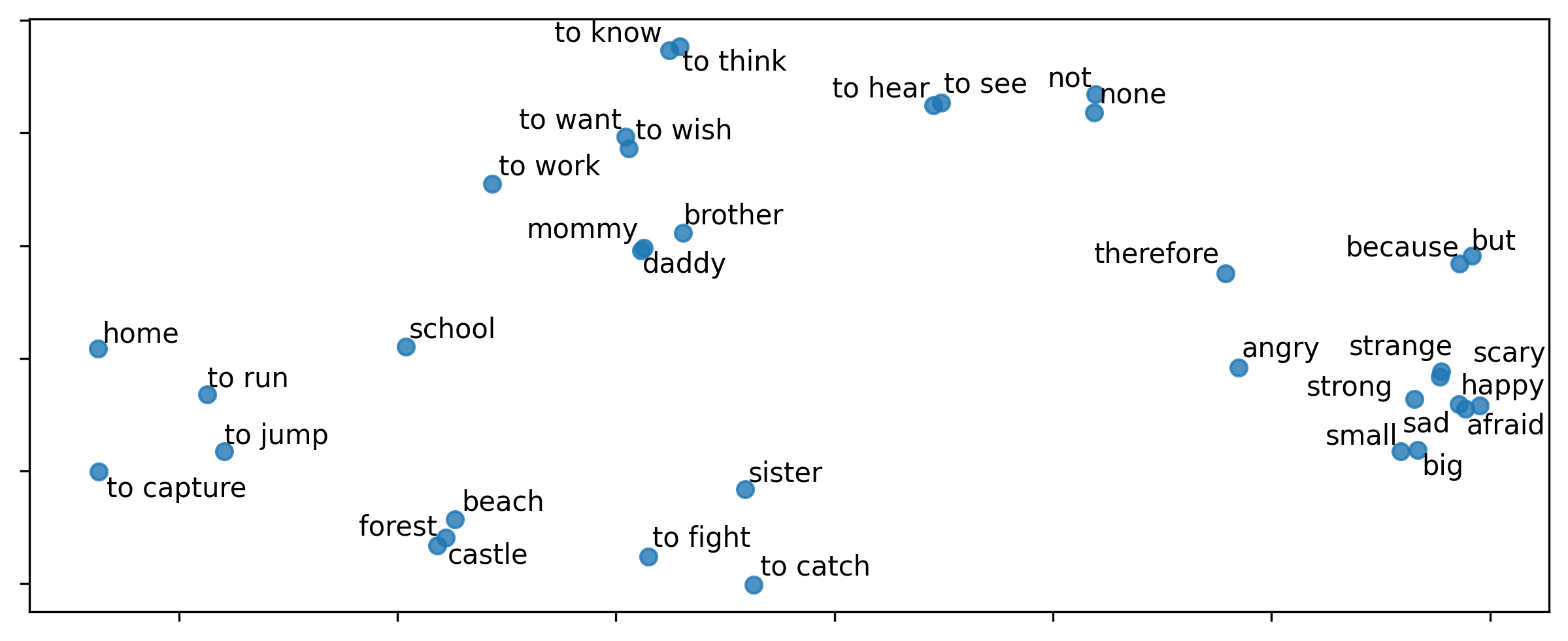}
    \end{subfigure}%
    \begin{subfigure}[b]{0.5\textwidth}
            \centering
            \includegraphics[width=\textwidth]{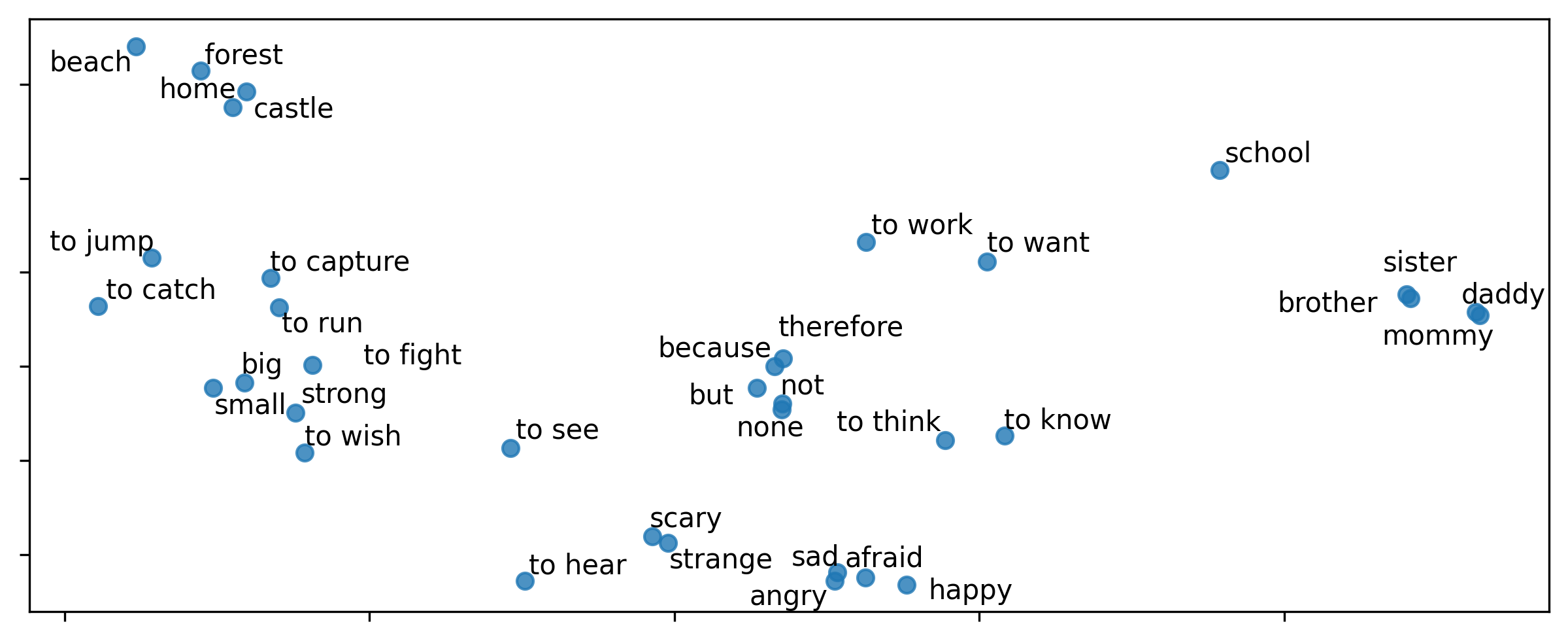}
    \end{subfigure}
    \caption{t-SNE projections \cite{JMLR:v9:vandermaaten08a} of the latent Word2Vec space of 100-dimensional lemma vectors of ChiSCor (left) and BasiScript (right). Lemma positions should not be compared \textit{between} but \textit{within} plots, as the axes of the plots have no explicit interpretation.}  
    \label{fig:w2v}
\end{figure*}

\subsection{Case study 3: Lexical Semantics with Word2Vec} \label{W2V}
The third case study demonstrates the usability of ChiSCor as a relatively small corpus with common NLP-tools. We use a Word2Vec model \citep{mikolov2013efficient} to visualize lexico-semantic differences in children's language use in ChiSCor and BasiScript. It is commonly assumed that training high quality word vectors requires large corpora (> 100 million tokens) \citep{mikolov2013efficient, altszyler2016comparative}; ChiSCor and BasiScript are much smaller with \textasciitilde74k and \textasciitilde3.4m tokens respectively. Still, it is worthwhile to see how well ChiSCor allows a computer to infer lexico-semantic information, since vector representations are the starting point for many downstream NLP tasks, and research in computational and cognitive linguistics \citep[e.g.][]{beekhuizen2021probing, samir-etal-2021-formidable}.

We obtained lemma vectors from both ChiSCor and BasiScript (introducced in Section \ref{Zipf}) with Word2Vec as implemented in Gensim 4.1.2 \citep{rehurek2011gensim}. For ChiSCor, the CBOW algorithm yielded the best result, for BasiScript this was Skip-gram. Vector quality was evaluated visually during training with reduced-dimensionality plots of a set of 35 common nouns, verbs, connectives, etc. that occur proportionally in both corpora. The end results are given in Figure \ref{fig:w2v}. Here we see that overall vectors from both corpora allow intuitive syntactic groupings (e.g. conjunctions `but'/`because', and verbs `to think'/`to know'), and semantic groupings (e.g. `mommy'/`daddy', `not'/`none'). To verify this quantitatively, we computed cosine similarities between the 595 possible pairs of the 35 lemmas plotted in Figure \ref{fig:w2v} with $\cos ({\bf v},{\bf w})= {{\bf v} \cdot {\bf w} \over \|{\bf v}\| \|{\bf w}\|}$, where ${\bf v}$ and ${\bf w}$ are two lemmas from one corpus, and computed their overlap. We found a fair correlation $\rho(595)=.45, p <.01$ \citep{akoglu2018user}, which is salient: it shows that from ChiSCor as relatively small corpus, rich lexico-semantic information can be learned as effectively as from BasiScript, which is 46 times larger.

Lemma vectors also allow us to analyze how children use particular lemmas of interest. There is some nuance in the groupings in Figure \ref{fig:w2v}: for ChiSCor, especially the verbs referring to cognitive states (`to think', `to know', `to wish', `to want') and perceptual states (`to hear', `to see') are more clearly grouped and positioned compared to BasiScript (where e.g. `to wish', `to see', and `to want' have less obvious positions). Since these lemmas have about equal relative frequencies in both corpora, it is likely that for these verbs, the lemma \textit{context} is semantically more clear and coherent in ChiSCor compared to BasiScript. On the other hand, conjunctions (`but', `because', `therefore') are more coherently grouped in BasiScript compared to ChiSCor (where `therefore' has a less obvious position). 

Apparently, children use verbs referring to cognitive/perceptual states more coherently in ChiSCor, while conjunctions are more coherently used in BasiScript. In live storytelling, communicating clearly and coherently what was thought and/or perceived seems more critical than in written storytelling, as the audience cannot access earlier information as they could in a written story, and this information is critical for understanding and relating to narratives more generally \citep{zunshine2006we}. On the other hand, in written stories, children have more time to reflect on, and, if necessary, correct their use of conjunctions to link clauses, making the context more clear and coherent. This example shows that ChiSCor is usable with common NLP-tools to unravel children's language use in detail, even though it is relatively small.

Lemma vectors can also reveal bias in children's speech. A well-known gender bias in language is the women-home/man-work stereotype \citep{bolukbasi2016man, wevers-2019-using}, which in ChiSCor and BasiScript can be investigated with gendered categories `mommy', and `daddy', and attributes `home' and `to work'. As we see in Figure \ref{fig:w2v}, `mommy' and `daddy' occupy similar positions, so initially we do not expect much difference in their cosine similarity with `home' and `to work'. A standard approach to verify this, is to compute the difference in cosine similarity of an attribute with one category versus another, e.g. `home' and `mommy' vs. `daddy'. For ChiSCor, difference scores were small: for `home' and `mommy' vs. `daddy' .031, for `to work' and `mommy' vs. `daddy' .076. The difference scores were comparably small for BasiScript: .049 and .001 respectively. These smaller scores indicate that neither gender is much more strongly associated with one attribute than the other, suggesting little gender bias in the corpora, contra earlier work on bias in child language \citep[e.g.][]{charlesworth2021gender}. Still, future work should leverage ChiSCor and incorporate more gendered categories (e.g. `she', `he'), more attributes (e.g. `baby', `office'), average these vectors and apply more advanced vector arithmetic to put this initially surprising result to the test.

\section{Discussion}\label{Discussion} 
Storytelling datasets are relatively scarce, which is a shortcoming in existing resources, given that live storytelling challenges children to leverage both linguistic, cognitive, and social competences to tell a story that engages an audience. These competences can be analysed through stories, manually or with computational tools, to learn more about child development. We demonstrated that ChiSCor has properties that other established language samples also have, such as a Zipfian token distribution. Moreover, ChiSCor's close fit to the Zipfian curve testifies to the \textit{social context} of the language contained in it and the Principle of Least Effort that is likely at work there (Section \ref{Zipf}). 

In addition, even though storytelling is a cognitively demanding task, we demonstrated that the stories in ChiSCor are syntactically surprisingly complex, and we offered a tentative explanation why especially younger children may employ complex syntax, which could be related to ChiSCor's context of live storytelling in front of an audience (Section \ref{Syntactic_comp}). Lastly, we have shown that ChiSCor can be used to learn a semantic vector space that is as intuitive as the semantic space of a much larger reference corpus (Section \ref{W2V}). This opens up possibilities for using ChiSCor with tools that are traditionally deemed fit only for much larger corpora, to assess the coherence of contexts in which children use particular words of interest. For example, we found that words detailing cognitive and perceptual states were more clearly differentiated in ChiSCor compared to BasiScript as a corpus of written child language. Such words concern information that is critical to understand a plot that cannot be consulted again in live storytelling, possibly leading children to use these words more carefully and coherently.

The social context of ChiSCor's narratives and its influence on language production invite us to reflect on a more general issue: the dominance of written (web) text in computational linguistics and NLP. Researchers increasingly question scraping together increasingly larger uncurated and undocumented resources \citep{bender2021dangers, PAULLADA2021100336}, that is, datasets without metadata, and it is subject to debate how helpful such large-scale written datasets are in e.g. understanding language acquisition and modelling cognition \citep[e.g.][]{warstadt2022artificial, mahowald2023dissociating}. Indeed, spoken language is different from written language in many ways, as \citet{linders2023zipfrevisited} note: it is mainly acquired naturally (unlike writing) and predates writing in both the evolutionary and developmental sense. Most critically, speech is typically situated in a social setting with other language users, evanescent, spontaneous, and grounded in a particular context, to mention just a few out of many defining characteristics.

Still, with Large Language Models (LLMs) as prime current example of the reliance on large written datasets, such datasets have helped disclose what is \textit{in principle} learnable from word co-occurrence statistics and a simple word prediction training objective, such as the capacity to represent language input hierarchically \citep{manning2020emergent}. Although we should take LLMs serious as the current best yet data-hungry distributional learners we have \citep{contreras2023large, dijk2023llm}, the next challenge is to achieve the same performance with more ecologically valid, smaller datasets and smaller neural architectures; here, corpora like ChiSCor could be part of the solution. Since ChiSCor has information on the age groups of the children who produced the language, future work could, for example, partition ChiSCor to employ train and/or test sets that more realistically model children's language use at different stages of their development. And since ChiSCor covers language from the speech domain, it provides an interesting opportunity to explore training language models on language with a different nature. Still, we do not mean to claim that ChiSCor solves all issues regarding LLMs and training data, but we hope to contribute a dataset that can be a part of the move towards better datasets for computational linguistics, a dataset that, in the words of \citet{bender2021dangers}, `is only as large as can be sufficiently documented'.         

Lastly, we like to emphasize that since ChiSCor features high-quality audio besides text, it naturally opens directions for multi-modal research. For example, research on detecting characters' emotions will benefit from adding information on prosody. Also, research aimed at improving speech-to-text models will benefit from the voices of 442 unique children of different ages, and accompanying transcripts, that can be used for fine-tuning existing speech-to-text models.

\section{Conclusion}\label{Conclusion}
This paper introduced ChiSCor as a versatile resource for computational work on the intersection of child language and cognition. ChiSCor is a new corpus of Dutch fantasy stories told freely by children aged 4-12 years, containing high-quality language samples that reflect the social settings in which they were recorded in many details. We provided three case studies as examples of how ChiSCor can fuel future work: studying language development with ChiSCor's out-of-the-box age metadata and linguistic features, modelling Zipf distributions with ChiSCor, and linking ChiSCor to common NLP-tools to study children's language in action. Besides verbatim and normalised texts, ChiSCor comes with 442 high-quality audio samples of 442 children, metadata on the backgrounds of 148 children, annotations of character complexity, and extracted linguistic features that will be useful for a variety of researchers. In addition to Dutch stories, ChiSCor comes with a small additional set of 62 English stories with the same additional metadata and annotations as for the Dutch stories.

Four years have passed since we started compiling ChiSCor. We look back on many great moments with the children who were happy to share their fantasies and cleverly constructed plots with us. We encourage readers of this paper to have a look at the corpus––both for research purposes and for fun.

\section*{Limitations}\label{lims}
Within the subset of our corpus that contains extra metadata (Section \ref{metadata},) older children and children from lower socioeconomic backgrounds are underrepresented. This may limit the generalizability of future work done with ChiSCor. This is partly due to a bias resulting from the way our metadata was obtained; the larger set of 619 stories is likely more balanced. A second limitation concerns character depth annotations: a large part of character depth labels depends on one expert. A third limitation is that for BasiScript, a license has to be signed before one can use it. Thus, we cannot provide its lexicon or the corpus on OSF, which makes parts of our study less directly reproducible. 

\section*{Ethics statement}\label{ethics}
In compiling this corpus, the researchers were frequently in touch with school principals, teachers, children and parents to find an appropriate way to collect, store and analyse the stories and metadata. Our study was reviewed and approved by the Leiden University Science Ethics Committee (ref. 2021-18). Regarding model efficiency, the spaCy models used to extract linguistic information are pre-trained, easy to use, and extraction of lexical and syntactic information did not take more than a couple of minutes. Further, the Gensim models used to train word vectors are also lightweight, easy-to-use, and equally efficient qua training time.

\section*{Acknowledgements}
This research was done in the context of Max van Duijn's research project `A Telling Story' (with project number VI.Veni.191C.051), which is financed by the Dutch Research Council (NWO). Besides the children, their teachers, and caregivers, we thank Isabelle Blok, Yasemin Tunbul, Nikita Ham, Iris Jansen, Werner de Valk, and Lola Vandame for help with data collection. We further thank Ageliki Nicolopoulou, Arie Verhagen, and Tom Breedveld for feedback on our annotations and analyses. We thank Li Kloostra for extensive comments on the final version of this paper. Lastly, we thank three anonymous reviewers for their constructive feedback.

\bibliography{anthology,custom}
\bibliographystyle{acl_natbib}

\end{document}